%% file: 11914_main.tex
\documentclass[10pt,twocolumn,letterpaper]{article}
\usepackage{graphicx}  
\usepackage{bm} 
\usepackage{multirow}
\usepackage{bbding}
\usepackage{cvpr}              
\usepackage[accsupp]{axessibility}

\input{preamble}

\definecolor{cvprblue}{rgb}{0.21,0.49,0.74}
\usepackage[pagebackref,breaklinks,colorlinks,citecolor=cvprblue]{hyperref}

\def\paperID{11914} 
\def\confName{CVPR}
\def\confYear{2024}

\title{Depth Information Assisted Collaborative Mutual Promotion Network for Single Image Dehazing}  

\author{Yafei Zhang\thanks{Yafei Zhang and Shen Zhou contribute equally to this work.}\ \ \ \ Shen Zhou$^*$\ \ \ \ Huafeng Li\thanks{Huafeng Li (lhfchina99@kust.edu.cn) is the corresponding author.}\\ 
Faculty of Information Engineering and Automation, Kunming University of Science and Technology\\
}

\begin{document}
\maketitle
\input{sec/0_abstract}    
\input{sec/1_intro}
\input{sec/2_relatedwork}
\input{sec/3_methods}

\input{sec/4_experiments}
{
    \small
    \bibliographystyle{ieeenat_fullname}
    \bibliography{main}
}


\end{document}

%% file: preamble.tex
\usepackage[dvipsnames]{xcolor}


%% file: sec/0_abstract.tex
\begin{abstract}
Recovering a clear image from a single hazy image is an open inverse problem. Although significant research progress has been made, most existing methods ignore the effect that downstream tasks play in promoting upstream dehazing. From the perspective of the haze generation mechanism, there is a potential relationship between the depth information of the scene and the hazy image. Based on this, we propose a dual-task collaborative mutual promotion framework to achieve the dehazing of a single image. This framework integrates depth estimation and dehazing by a dual-task interaction mechanism and achieves mutual enhancement of their performance. To realize the joint optimization of the two tasks, an alternative implementation mechanism with the difference perception is developed. On the one hand, the difference perception between the depth maps of the dehazing result and the ideal image is proposed to promote the dehazing network to pay attention to the non-ideal areas of the dehazing. On the other hand, by improving the depth estimation performance in the difficult-to-recover areas of the hazy image, the dehazing network can explicitly use the depth information of the hazy image to assist the clear image recovery. To promote the depth estimation, we propose to use the difference between the dehazed image and the ground truth to guide the depth estimation network to focus on the dehazed unideal areas. It allows dehazing and depth estimation to leverage their strengths in a mutually reinforcing manner. Experimental results show that the proposed method can achieve better performance than that of the state-of-the-art approaches. The source code is released at \href{https://github.com/zhoushen1/DCMPNet}{\textcolor{blue}{https://github.com/zhoushen1/DCMPNet}}. \vspace{-3mm}
\end{abstract}

%% file: sec/1_intro.tex
\begin{figure}[t]  
	\centering
	\includegraphics[width=1.0\linewidth]{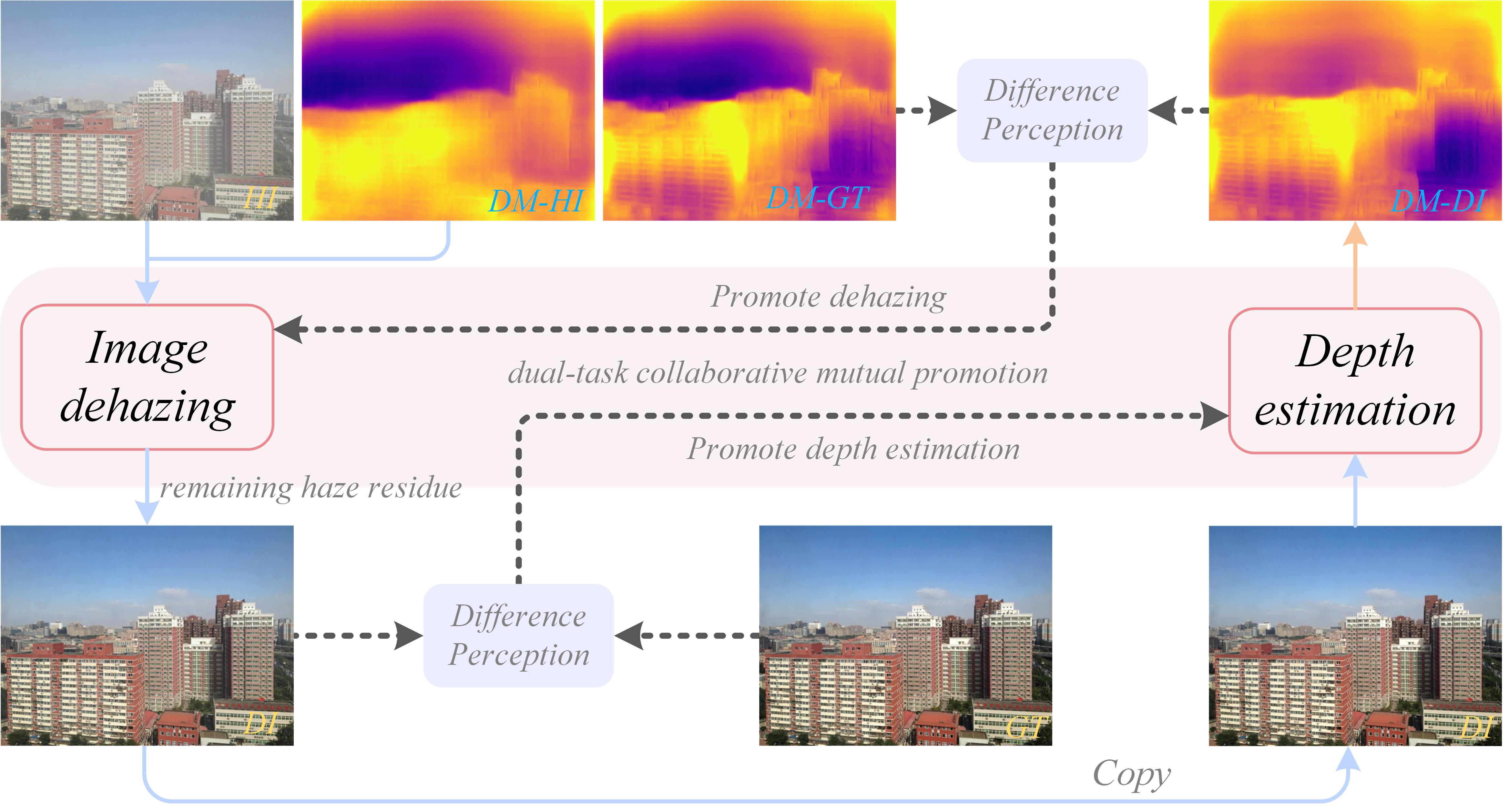}
	\caption{Idea of dual task collaboration and mutual promotion. HI denotes the hazy image. \textit{DM-HI}, \textit{DM-GT} and \textit{DM-DI} denote the depth maps of hazy image, ground truth and dehazed image, respectively.} \vspace{-3mm}
	\label{Fig1}
\end{figure}
\section{Introduction}
\label{sec:intro}
\hspace*{\parindent}Single image dehazing refers to restoring a clear image from a given hazy image. This technology has attracted wide attention due to its critical role in downstream computer vision tasks \cite{sun2022rethinking,li2020netnet,huang2020dsnet,Xu2023Video,Kulkarni2023Aerial}. Mathematically, the atomization process is commonly simulated by employing the atmospheric scattering model \cite{mccartney1976optics,narasimhan2000chromatic,narasimhan2002vision}:
\begin{equation}
	\bm I(x) = \bm J(x) \bm T(x) + (1 - \bm T(x))\bm A(x)
	\label{eq:important}
\end{equation}
where $x$ represents the pixel position, $\bm I(x)$ denotes the hazy image, $\bm J(x)$ denotes the clear image, $\bm T(x)$ indicates the transmission map, and $\bm A(x)$ represents the atmospheric light. The transmission map $\bm T(x)$ can also be expressed as:
\begin{equation}
	\bm T(x) = {e^{-\beta \bm d(x)}}
	\label{eq:important}
\end{equation}
where $\beta $ is the scattering coefficient, and $\bm d(x)$ is the scene depth.

From Eq.(1), we can see that image dehazing methods based on the atmospheric scattering model necessitate the estimation of both $\bm T(x)$ and $\bm A(x)$. However, most studies claim that $\bm T(x)$ primarily contributes to generating haze within images. Consequently, the most existing Eq.(1)-based methods focus on estimating the transmission map $\bm T(x)$. For $\bm A(x)$, the maximum pixel value is taken as its value \cite{tan2008visibility}. In practice, the maximum pixel value in an image may originate from the brightest object rather than representing atmospheric light. To address this issue, the methods in \cite{berman2017air,sulami2014automatic,li2022haze} consider the estimation of both $\bm T(x)$ and $\bm A(x)$, achieving an improvement in dehazing performance. However, such atmospheric scattering model-based methods significantly depend on $\bm T(x)$ and $\bm A(x)$. In reality, the uneven distribution of haze poses a challenge in accurately estimating $\bm T(x)$ in each local region of the image, which limits the improvement of dehazing performance.

End-to-end image dehazing methods recover clear images directly from hazy images without the aid of atmospheric scattering models \cite{guo2022image,dong2020multi,wu2021contrastive,qin2020ffa,liu2019griddehazenet}, thereby getting rid of the dependence on $\bm T(x)$ and $\bm A(x)$. However, the absence of guidance from the atmospheric scattering model also poses challenges for restoring clear images. To solve this problem, image dehazing based on prior information was proposed \cite{Berman2016NonlocalID,chen2021psd,liu2023nighthazeformer,wu2023ridcp}. Specifically, PSD \cite{chen2021psd} establishes a loss committee consisting of dark channel prior \cite{he2010DCP}, bright channel prior \cite{zhang2021BCP} and histogram equalization to guide the restoration of clear images. NHFormer \cite{liu2023nighthazeformer} uses dark channel priors and bright channel priors to guide the generalization of the dehazing model from synthetic domains to real-world applications. RIDCP \cite{wu2023ridcp} obtains high-quality codebook priors through pre-trained VQGAN. These priors are used to facilitate controllable high-quality priors matching, thereby achieving feature recovery for hazy images. In addition, through the two-stage network structure design, MITNet \cite{shen2023MITNet} simultaneously realizes the joint recovery of features in both the time and frequency domains and achieves the restoration of both amplitude and phase spectrum, ensuring the quality of dehazed images.

Although the above methods are effective, they ignore the correlation between the depth information of the hazy image and itself. According to Eq.(2), the depth information $\bm d(x)$ of the scene is directly related to $\bm T(x)$, while $\bm T(x)$ is one of the key factors that lead to the appearance of haze in image. Suppose $\bm d(x)$ can be accurately estimated from its hazy image, and the calculated result can be used to reconstruct the hazy-free image. In that case, it will help improve the performance of the dehazing model. Based on this idea, this paper proposes a dual-task collaborative mutual promotion network for single image dehazing within the framework of end-to-end deep learning. In the proposed method, we embed depth estimation and image dehazing within a unified framework. To effectively use the depth information of the hazy image for the dehazing model, a difference perception-based dual-task interaction mechanism is designed to serve as a bridge between the two tasks. This mechanism seamlessly integrates depth estimation and image dehazing to form a dual-task-driven dehazing method.

The main idea of the proposed approach is shown in Figure \ref{Fig1}. The proposed method improves the depth estimation on hazy images by perceiving the difference between the output results of the dehazing network and the expected results so that the dehazing network can receive high-quality depth estimation information as guidance in the dehazing process. Moreover, the design of the dual-task collaborative mutual promotion framework will be conducive to the dehazing network to learn optimal network parameters. In terms of depth estimation improving the dehazing performance, it makes the dehazing network pay attention to areas where the dehazing effect could be unsatisfactory by perceiving the depth information difference between the dehazing result and the ideal image. In terms of the dehazing network promoting depth estimation, the robustness of the depth estimation is improved by making the depth estimation network pay attention to the non-ideal dehazing areas and obtaining more accurate prediction results on the hazy image. We have verified the effectiveness of the proposed method on three dehazing benchmarks and achieved compatible results of the state-of-the-art.

%% file: sec/2_relatedwork.tex
\section{Related Work }
\label{sec:relatedwork}

\subsection{Model-based Image Dehazing}
\hspace*{\parindent}Single image dehazing based on the atmospheric scattering model is a popular method. In this kind of method, the accuracy of the transmission map and atmospheric light estimation are pivotal factors affecting the quality of the dehazing results. To this end, a large number of dehazing methods based on transmission map estimation have emerged \cite{cai2016dehazenet,ren2016single,wang2020abc,pang2020bidnet}. 
These methods ignore the influence of atmospheric light on the dehazing results. To solve this problem, a joint estimation model for transmission map and atmospheric light is proposed. Particularly, Zhang \textit{et al}. \cite{zhang2018densely} proposed a joint discriminator based on GAN to assess the accuracy of the dehazed image and the estimated transmission map. Additionally, they employed a U-Net to predict the atmospheric light. Li \textit{et al}. \cite{Li2019ICCV} proposed a multi-stage progressive learning approach to estimate the transmission map. Additionally, they used a global average pooling layer to normalize the features from different layers and superimpose them together, and thus achieved estimating of the atmospheric light. Guo \textit{et al}. \cite{guo2019dense} used a shared DensetNet encoder and two non-shared DensetNet decoders to estimate the transmission map and atmospheric light jointly. 
Lee \textit{et al}. \cite{lee2020cnn} proposed a feature extraction network with a joint constraint of transmission map, atmospheric light and dehazing result. The network was used to predict more comprehensive information in the transmission map and atmospheric light to improve the quality of dehazing results. 

Due to the absence of prior information, the dehazing results from the above methods still need improvement. To this end, prior knowledge was introduced to estimate transmission maps and atmospheric light \cite{liu2019learningA,liu2019learningD,zhao2021refinednet,fan2023non,zhou2023physical}. Commonly used prior knowledge includes dark channel prior \cite{he2010DCP}, bright channel prior \cite{zhang2021BCP} and color attenuation prior \cite{zhu2015Color}, etc.
Although the above methods have proven effective, their performance depends on the estimation accuracy of the transmission map and atmospheric light. Compared with the physical model-driven approaches, end-to-end dehazing methods based on deep learning have attracted researchers' attention because they eliminate the constraints of transmission maps and atmospheric light.

\subsection{End-to-End Image Dehazing}
\hspace*{\parindent}End-to-end dehazing methods usually restore a clear image directly from the hazy image. To improve the visual quality of dehazing results, Ren \textit{et al}. \cite{ren2018gated} used white balance, image contrast enhancement and gamma transformation to preprocess the original image. They then determined the weights of the three processed images using a gating mechanism, ultimately achieving the restoration of hazy images. Wu \textit{et al}. \cite{wu2021contrastive} proposed a regularization method based on contrast learning, which achieved the recovery of a clean image by pushing away the distance between the dehazing result and the original image and simultaneously narrowing the distance between the dehazing result and the label image. Given its proficiency in highlighting useful information, attention-based techniques have been widely used in image dehazing \cite{liu2019griddehazenet,qin2020ffa}. Moreover, 
Guo \textit{et al.} \cite{guo2022image} proposed an image dehazing network combining CNN and Transformer, which leveraged the complementary characteristics of CNN and Transformer in feature extraction. Due to the high computational complexity of softmax-attention in transformer, Qiu \textit{et al}. \cite{Qiu2023ICCV} proposed to approximate it with Taylor expansion, effectively reducing the complexity of traditional attention mechanisms in image dehazing. The above methods only consider the information recovery from the time domain, ignoring the influence of hazy information on the frequency domain. Therefore, Shen \textit{et al}. \cite{shen2023MITNet} proposed to restore hazy-free image information in both the time and frequency domain.

Although the end-to-end dehazing methods do not rely on the physical model of the hazy image, they exhibit a high dependency on the training dataset. If we find a more effective way to leverage the information in the training samples, it will help improve the model performance. The distribution of haze has a relationship with the scene depth, Yang \textit{et al}. \cite{yang2022depth} proposed a depth perception method to estimate the depth map of the hazy image and provide depth features for dehazing in a unified framework. Considering the challenge posed by the absence of ground truth of hazy images in real scenarios, Yang \textit{et al}. \cite{yang2022_D4} proposed a paired sample construction method for synthesizing authentic hazy images from clear haze-free images, thereby realizing the supervised training of the dehazing model. Wang \textit{et al}. \cite{wang2023selfpromer} proposed a self-improving depth-consistent dehazing network according to the differences in depth features between hazy image and its ground truth. Nevertheless, the method fails to achieve collaborative mutual promotion between depth estimation and image dehazing, which limits the model performance. In this work, we fully consider the beneficial impact of depth information in hazy images on dehazing and propose to treat depth estimation and image dehazing as two independent tasks. The two tasks are seamlessly integrated into a unified learning framework through a dual-task interaction mechanism, which realizes collaborative mutual promotion.

%% file: sec/3_methods.tex
\section{The Proposed Method}
\label{sec:methods}
\begin{figure}[t!]
	\centering
	\includegraphics[width=1.0\linewidth]{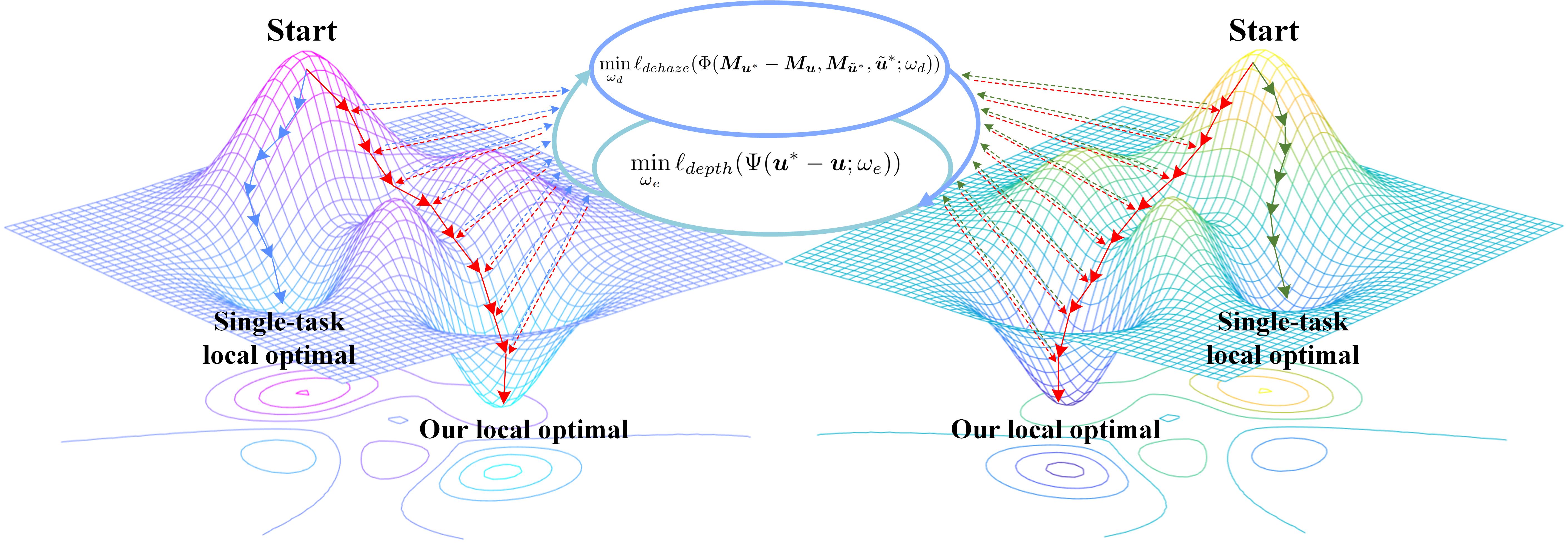}
	\caption{Dual-task Collaborative optimization formulation for image dehazing and depth estimation.}\label{Fig2} \vspace{-3mm}
\end{figure}

\begin{figure*}[t!]
	\centering
	\includegraphics[width=0.98\linewidth]{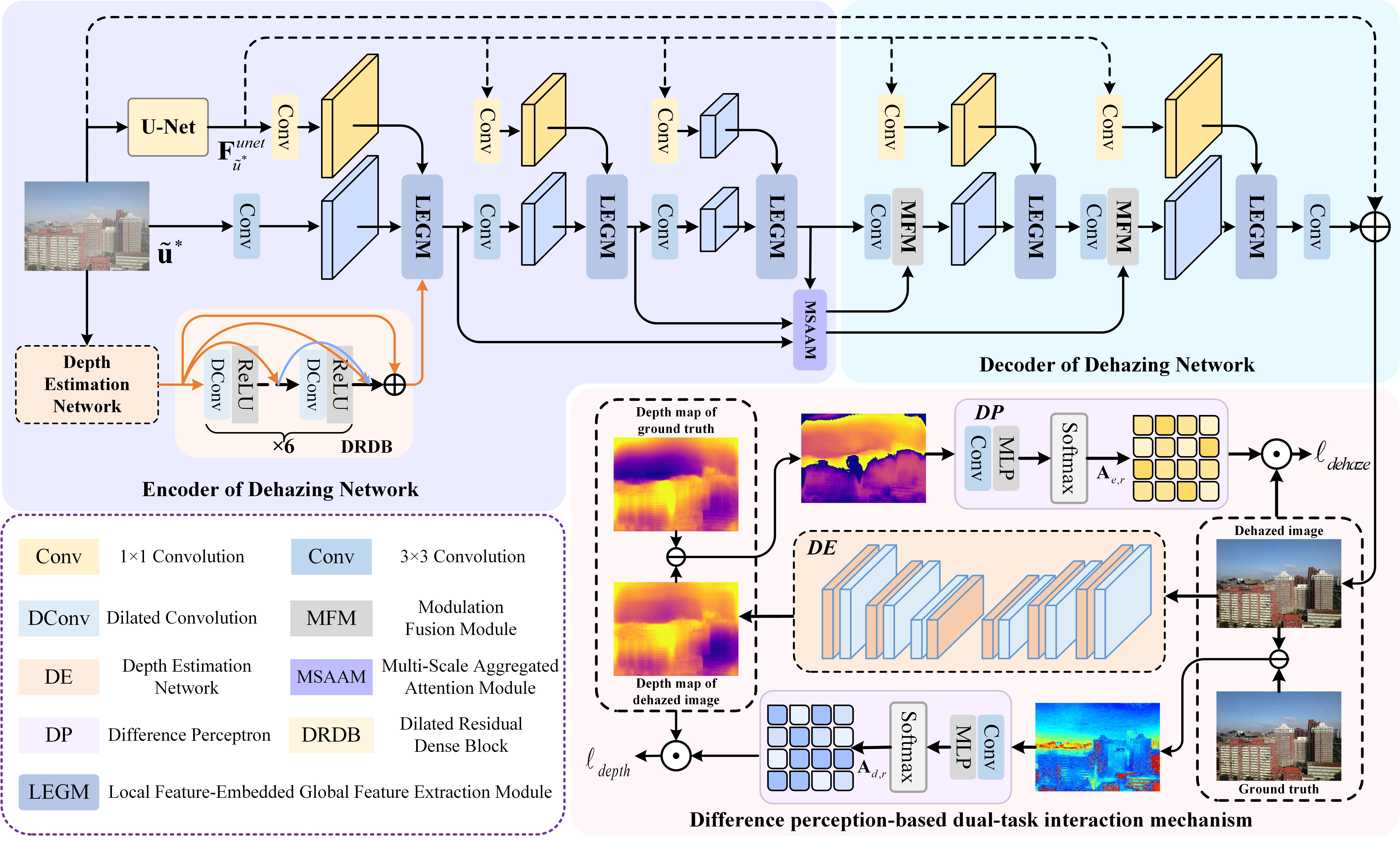}
	\caption{Architecture of the proposed method, which consists of the dehazing network, depth estimation network (DE) and difference perception (DP). DE and DP are the main components of difference perception-based dual-task interaction mechanism. This mechanism enables dehazing and depth estimation to be seamlessly integrated into a unified framework, and improves the model performance of two tasks through collaborative mutual promotion.}
	\label{Fig3} \vspace{-3mm}
\end{figure*}

\subsection{Idea formulation}
\hspace*{\parindent}In this paper, we propose a single image dehazing method assisted by depth information. Unlike existing image dehazing methods that rely on elaborately designed algorithms to yield satisfactory results, our method improves the dehazing performance by leveraging auxiliary task sensitive to the quality of dehazing results. Assuming that the hazy image is $\tilde{\bm u}^ *$, the dehazed image is ${\bm u}^*$, and the ground truth (GT) is $\bm u$, the idea of depth estimation based on the dehazed image can be formulated as follows:
\begin{equation}
	\mathop {\min }\limits_{{\omega _e}} {\ell_{depth}}(\Psi({\bm u}^*-\bm u; \omega_e )),
\end{equation}
where $\Psi$ is the depth estimation network with parameter $\omega_e$, $\ell_{depth}$ is specific depth estimation loss. As demonstrated in Eq.(3), the difference between ${\bm u}^*$ and $\bm u$ is used to optimize the parameter $\omega_e$. This enables $\Psi$ to improve its performance on ${\bm u}^*$ by focusing on the differences between ${\bm u}^*$ and $\bm u$. Since one of the contributors to the difference between ${\bm u}^*$ and $\bm u$ stems from the presence of residual haze information and regions where the dehazing effect is non-ideal. If the depth estimation network can improve its performance in these regions, the dehazing result can be also improved in the corresponding regions of $\tilde{\bm u}^*$. 

For optimizing parameter $\omega_d$ of $\Phi $, this paper employs the following optimization objective:
\begin{equation}
	\mathop {\min}\limits_{{\omega_d}} {\ell_{dehaze}}(\Phi ( {\bm M}_{{\bm u}^*}-{\bm M}_{\bm u}, {\bm M}_{\tilde{\bm u}^*}, \tilde{\bm u}^*; \omega _d)),
\end{equation}
where $\ell_{dehaze}$ is the specific dehazing loss, $\Phi$ is the dehazing network with parameter $\omega _d$, ${\bm M}_{{\bm u}^*}$, ${\bm M}_{\bm u}$ and ${\bm M}_{\tilde{\bm u}^*}$ are the depth maps of ${\bm u}^*$, $\bm u$ and $\tilde{\bm u}^*$ respectively. In this work, RA-Depth \cite{he2022RA} is used to generate ${\bm M}_{\bm u}$. In Eq.(4), ${\bm M}_{{\bm u}^*}-{\bm M}_{\bm u}$ is the difference between the depth maps of ${\bm u}^*$ and $\bm u$. Suppose this difference is fed back into the training process of dehazing network $\Phi$, and used to prompt $\Phi$ to focus on areas where depth estimation is not ideal. In that case, the performance of $\Phi $ will be improved by Eq.(4). Driven by this idea, the depth estimation network $\Psi$ will play a positive role in promoting the performance of the dehazing network $\Phi$. Furthermore, as the performance of $\Psi$ on $\tilde{\bm u}^*$ improves, the predicted depth information becomes more accurate, leading to a more comprehensive representation of structural details in the depth map. Consequently, injecting ${\bm M}_{\tilde{\bm u}^*}$ into $\Phi$ will have a beneficial effect on the recovery of structural information in hazy images. This also reflects the promotion effect of $\Psi$ on $\Phi$. As shown in Figure \ref{Fig2}, this design idea of dual-task collaboration can also effectively prevent  the dehazing network parameter $\omega_d$ from falling into an undesirable local optimum \cite{liu2023multi}. 

\subsection{Depth Information Assisted Image Dehazing}
\subsubsection{Encoder of Dehazing Network}
\begin{figure}[htbp]
	\centering
	\includegraphics[width=1.0\linewidth]{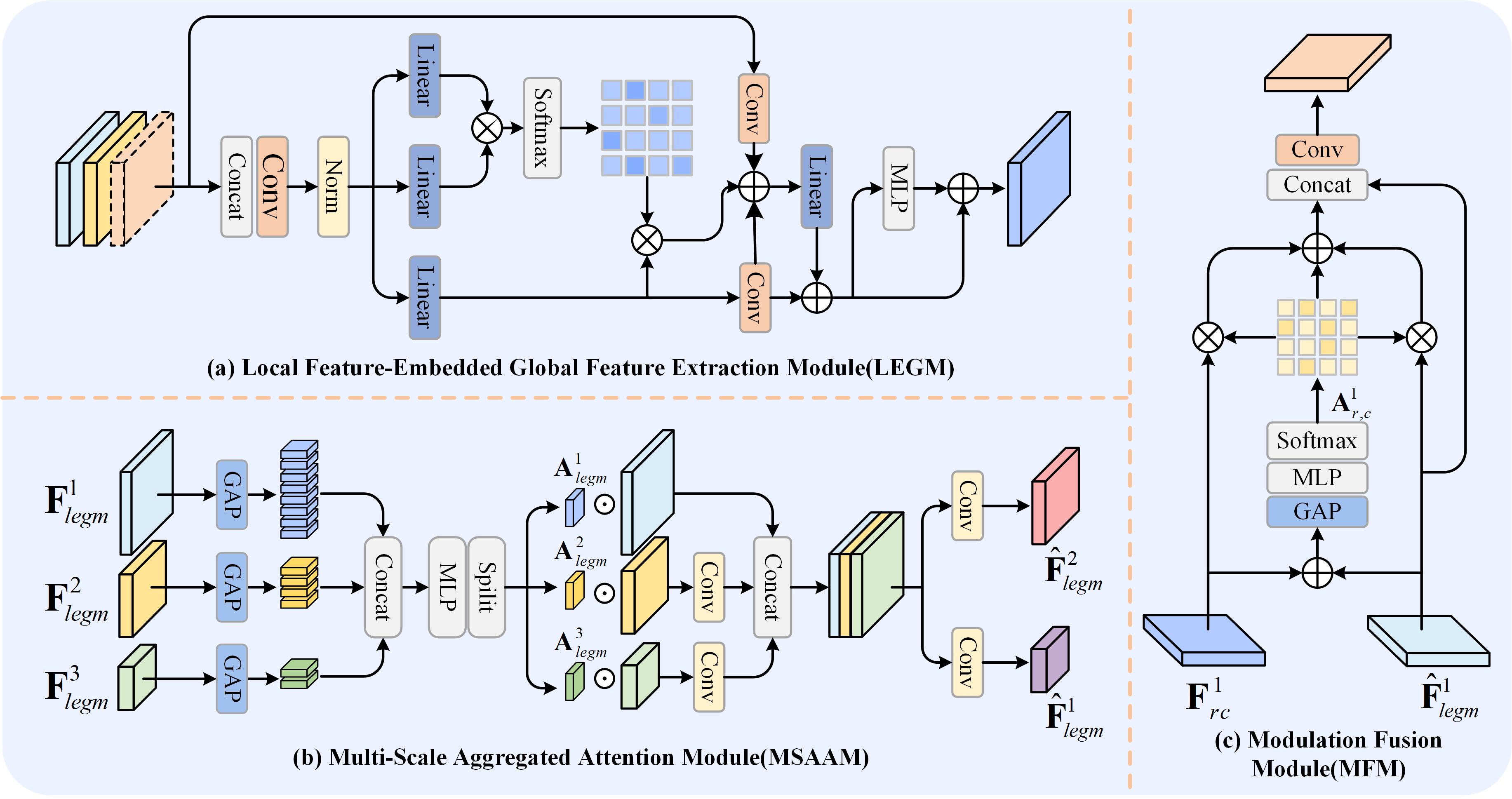}
	\caption{Architectures of the LEGM, MSAAM, MFM.}\label{Fig4} \vspace{-3mm}
\end{figure}

\hspace*{\parindent}As seen from Figure \ref{Fig3}, the depth information-assisted image dehazing network consists of encoder and decoder. The encoder is mainly composed of U-Net, local feature-embedded global feature extraction module (LEGM), depth estimation network (DE), dilated residual dense block (DRDB) and multi-scale aggregation attention module (MSAAM). To ensure the quality of U-Net output, the parameters of U-Net are optimized by $l_{1}$-loss:
\begin{equation}
	\ell_{unet} = ||\bm F_{\tilde{\bm u}^*}^{unet} -\bm F_{{\bm u}}^{unet}||_1
\end{equation}
where $\bm F_{\tilde{\bm u}^*}^{unet}$ and $\bm F_{{\bm u}}^{unet}$ denote the features of $\tilde{\bm u}^*$ and $\bm u$ of U-Net output respectively.  As shown in Figure \ref{Fig4} (a), the self-attention block is the main component of LEGM. Its inputs include the features output by the $1\times1$ convolution after the U-Net, the features output by the $3\times3$ convolution and the features output by the DRDB after the DE. Since the features extracted by the convolutional network contain a large number of local information, we name the self-attention block combined with the convolution layer LEGM. In the depth information-assisted dehazing, only the first LEGM receives the depth information of the hazy image. The encoder of the dehazing network contains three LEGMs, and the outputs of these LEGMs are integrated through the MSAAM shown in Figure \ref{Fig4} (b) to prevent the loss of shallow features.

Let the outputs of three LEGMs be $\bm F_{legm}^1$, $\bm F_{legm}^2$ and $\bm F_{legm}^3$ respectively. The three outputs are then processed through GAP, Concat, MLP and split operations, the weight matrices $\bm A_{legm}^1$, $\bm A_{legm}^2$ and $\bm A_{legm}^3$ are obtained and used for modulating them. The modulated features can be expressed as:
\begin{equation}
	\tilde{\bm F}_{legm}^i = \bm A_{legm}^i \odot \bm F_{legm}^i \quad (i = 1,2,3)
\end{equation}
where $\odot$ denotes dot product. After modulation by Eq. (6), the features related to the clear image in $\bm F_{legm}^i$ $(i = 1,2,3)$ will be highlighted, which is conducive to the protection of detailed information. Since $\tilde{\bm F}_{legm}^1$, $\tilde{\bm F}_{legm}^2$ and $\tilde{\bm F}_{legm}^3$ have different dimensions, to mine the potential relationship between them, the convolution operation is respectively employed on $\tilde{\bm F}_{legm}^2$ and $\tilde{\bm F}_{legm}^3$ to facilitate their dimensional alignment with $\tilde{\bm F}_{legm}^1$. Subsequently, $\tilde{\bm F}_{legm}^1$, $\tilde{\bm F}_{legm}^2$ and $\tilde{\bm F}_{legm}^3$ are concatenated and processed through two convolutional layers with non-shared parameters, yielding the features $\hat{\bm F}_{legm}^1$ and $\hat{\bm F}_{legm}^2$. 

\subsubsection{Decoder of Dehazing Network}
\hspace*{\parindent}As can be seen from Figure \ref{Fig3}, the decoder of the dehazed image mainly consists of two LEGMs with feature modulation implantation (FMI). Each of them consists of two convolution layers, a modulation fusion module (MFM) and an LEGM, where the structure of the MFM is shown in Figure \ref{Fig4} (c). The inputs of the first MFM are $\hat{\bm F}_{legm}^1$ and the feature $\bm F_{rc}^1$ derived after $\tilde{\bm F}_{legm}^3$ undergoes $3\times 3$ convolution. The result obtained after the sum of $\hat{\bm F}_{legm}^1$ and $\bm F_{rc}^1$ is processed through GAP, MLP, and Softmax, is denoted as $\bm A_{r,c}^1$. The values in $\bm A_{r,c}^1$ indicate the significance of the features in $\hat{\bm F}_{legm}^1$ and $\bm F_{rc}^1$ in the reconstruction of the dehazed image. Their contribution to the image dehazing can be highlighted by adjusting with $\bm A_{r,c}^1$. The specific process can be expressed as follows:
\begin{equation}
	\tilde{\bm F}_{rc}^1 = \bm A_{r,c}^1 \odot \hat{\bm F}_{legm}^1 + \bm A_{r,c}^1 \odot \bm F_{rc}^1
\end{equation}
$\tilde{\bm F}_{rc}^1$ and $\hat{\bm F}_{legm}^1$ are concatenated to enhance the shared information between them. Subsequently, the concatenation result is processed through a convolution layer to yield the output of the first LEGM with FMI. In the second LEGM with FMI, the first output of LEGM with FMI and $\hat{\bm F}_{legm}^2$ are used as its inputs, and the final output is sent to a convolution layer to reconstruct the final dehazed image ${\bm u}^*$.

\subsection{Depth Estimation Driven by Difference Perception}
\hspace*{\parindent}The proposed dehazing method not only addresses the image dehazing but also incorporates depth estimation based on the dehazed images, thereby presenting a dual-task collaborative promotion network framework. Regarding network architecture, DE follows a U-Net structure, adopting a DRDB-based encoder and decoder design. The encoder and decoder comprises $4$ DRDBs, respectively. To improve the performance of the DE on hazy images, we introduce the difference perception between dehazing images and their GTs. The process is shown in the difference perception-based dual-task interaction of Figure \ref{Fig3}. 

The difference between the dehazing result ${\bm u}^*$  and $\bm u$ is:
\begin{equation}
	\bm R_{{\bm u}^*/{\bm u}} = {\bm u}^*-\bm u
\end{equation}
where $\bm R_{{\bm u}^*/{\bm u}}$ reflects the difference between the dehazing result and the label. It also indirectly indicates the location of regions in the dehazing result where the dehazing effect is not ideal. Suppose the DE can focus on those regions and yield accurate depth estimation results in these regions. In that case, it will provide more accurate depth information for image dehazing in the mutual promotion of dual tasks to assist the dehazing network in obtaining ideal dehazing results. In this process, the difference $\bm R_{{\bm u}^*/{\bm u}}$ is processed through a difference perceptron consisting of convolutional layers, MLP and Softmax to predict the coefficient matrix $\bm A_{d,r}$ that reflects the difference between ${\bm u}^*$ and $\bm u$. DE is optimized by minimizing a tailored loss function as follows:
\begin{small}
\begin{equation}
	\mathop {\ell_{depth}} = {\left\| \bm A_{d,r} \odot (\bm M_{{\bm u}^*} - \bm M_{\bm u}) \right\|_1} + {\left\| \bm M_{\tilde{\bm u}^*} -\bm M_{\bm u} \right\|_1}
	\label{eq:important} \vspace{-3mm}
\end{equation} 
\end{small}

It is worth noting that optimizing the DE guided by Eq. (9) embodies an integrated optimization strategy of dual tasks. On the one hand, the DE ensures consistency between the estimated depth maps of dehazed images and their corresponding GTs, particularly in regions that exhibit suboptimal dehazing results. This promotes the DE to allocate increased attention to the regions where the dehazed image differs from the GT. The difference is mainly from the residual haze in the dehazed image. Therefore, the optimization in Eq. (9) can improve the dehazing network's perceptiveness to the haze information in the hazy image. On the other hand, the constraint on hazy image depth estimation is introduced into the loss function in Eq. (9), which is equivalent to optimizing the DE via a dual-task strategy. As depicted in Figure \ref{Fig2}, this strategy can effectively mitigate the limitation of a single-task DE that may converge to a non-ideal local optimal solution, enhancing overall performance robustness. 

\subsection{Dual-task Collaborative Mutual Promotion Learning}
\hspace*{\parindent}DE and the image dehazing network are trained in a mutually promoting manner. When optimizing the DE, the main loss function is ${\ell_{depth}}$. Since $\bm M_{{\bm u}^*} = \Psi({\bm u}^*, \omega_e)$, $\bm M_{\tilde{\bm u}^*} = \Psi(\tilde{\bm u}^*, \omega_e)$ and ${\bm u}^* = \Phi(\tilde{\bm u}^*, \omega_d)$, 
we have $\bm M_{{\bm u}^*}=\Psi(\Phi(\tilde{\bm u}^*, \omega_d), \omega_e)$, and then the formula for updating parameter $\omega_e$ can be expressed as:
\begin{equation}
	\begin{aligned}
		&\mathop {\omega_e} \leftarrow {\omega_e} -\\& {\eta_e}{\nabla_{{\omega_e}}}{\ell_{depth}}(\Psi(\Phi(\tilde{\bm u}^*, \omega_d),\omega_e); \Psi(\tilde{\bm u}^*,\omega_e); \bm M_{\bm u})
	\end{aligned}
\end{equation}
where ${\eta_e}$ denotes the learning rate. $\Psi(\Phi(\tilde{\bm u}^*, \omega_d),\omega_e)$, $\Psi(\tilde{\bm u}^*,\omega_e)$ and $\bm M_{\bm u}$ represent the inputs of the loss function ${\ell _{depth}}$. It can be seen that the output result ${\bm u}^* = \Phi(\tilde{\bm u}^*, \omega_d)$ is used when updating the parameter $\omega_e$. In the iterative optimization of $\Phi$ and $\Psi$, the performance of the dehazing network $\Phi$ increases from weak to strong. Under the constraint of loss function ${\left\| \bm A_{d,r} \odot (\bm M_{{\bm u}^*}-\bm M_{\bm u}) \right\|_1}$ in Eq. (9), the DE can improve its accuracy by focusing on the difference between the dehazing result and the GT. This demonstrates that the dehazing network plays a positive role in promoting the performance of the DE. 


During the optimization of the dehazing network, $\bm M_{{\bm u}^*}-\bm M_{\bm u}$ is fed to the difference perceptron to yield the coefficient matrix $\bm A_{e,r}$. With $\bm A_{e,r}$, we use the following loss function to update the parameters of the dehazing network:
\begin{equation}
	\begin{aligned}
		\mathop {\ell_{dehaze}} & =  {\left\| \bm A_{e,r} \odot ({\bm u}^*-\bm u) \right\|_1}\\		
		& + \sum\limits_{i = 1}^n {{\lambda _i}\frac{\left\| VGG_i(\bm u)-VGG_i(\Phi(\tilde{\bm u}^*, \omega_d))\right\|_1}{\left\| VGG_i(\tilde{\bm u}^*)-VGG_i(\Phi(\tilde{\bm u}^*, \omega_d))\right\|_1}}
	\end{aligned}
\end{equation}
where $VGG_i$ represents the output feature of the $i$-th layer of VGG19 \cite{simonyan2014VGG19}, and ${\lambda_i}$ is the weight coefficient \cite{wu2021contrastive}. Therefore, the update process of the dehazing network can be described by:
\begin{small}
\begin{equation}
	\begin{aligned}
		&\mathop {\omega_d} \leftarrow {\omega_d}-\\&{\eta _d}{\nabla_{{\omega_d}}}{\ell _{dehaze}}(\Phi(\tilde{\bm u}^*, \omega_d);\Psi(\Phi(\tilde{\bm u}^*, \omega_d),\omega_e); \bm u; \bm M_{\bm u})
	\end{aligned}
\end{equation}
\end{small}
where ${\eta _d}$ is the learning rate. It can be seen that when updating ${\omega _d}$, the depth estimation result of the dehazing image $\Psi(\Phi(\tilde{\bm u}^*, \omega_d),\omega_e)$ is used. In this process, the difference between $\Psi(\Phi(\tilde{\bm u}^*, \omega_d),\omega_e)$ and $\bm M_{\bm u}$ is directly conveyed to the optimization of ${\omega _d}$ through the loss function, which promotes the dehazing network to improve its performance by updating $\omega_d$, to reduce the difference between ${\bm u}^*$ and ${\bm u}$. Therefore, it is evident from the above analysis that the DE plays a positive role in promoting the performance of the dehazing network. On the other hand, as illustrated in Figure \ref{Fig3}, the depth estimation result of $\tilde{\bm u}^*$ is fed into the LEGM and participates in the feature extraction of hazy images. This facilitates the dehazing network to obtain useful auxiliary information from $\bm M_{\tilde{\bm u}^*}$ to improve the quality of the dehazing results. 

%% file: sec/4_experiments.tex
\section{Experiments}
\label{sec:experiments}

\subsection{Experimental Settings}

\paragraph{Datasets} To ensure an unbiased comparison with current dehazing methods, we use the Indoor Training Set (ITS) and Outdoor Training Set (OTS) from the RESIDE dataset \cite{li2018benchmarking} as the training data. For evaluation, we employ the Synthetic Objective Test Set (SOTS), which contains $500$ indoor and $500$ outdoor hazy images for testing. The real-world images are collected to verify the model generalization. \vspace{-4mm}

\paragraph{Implementation Details.} All experiments are conducted on an NVIDIA GeForce RTX 3090 with 24GB GPU, and the model is implemented in the Pytorch 1.12.0 framework. During the training phase, Adam optimizer \cite{Adam} 
is used to optimize the network. We set the initial learning rate to $0.001$ and use a cosine annealing strategy to adjust the learning rate. Moreover, we randomly crop the images into $256\times256$ patches for training. In each mini-batch, the patches are augmented through horizontal or vertical flipping to enlarge the training samples. The entire training process lasted for a total of 600 epochs on the indoor dataset and 60 epochs on the outdoor dataset. \vspace{-4mm}
\begin{table*}[htbp]\small
	\centering
	\caption{Performance of the proposed method is compared with that of state-of-the-art methods on the synthetic datasets (SOTS-indoor and SOTS-outdoor). Best values are in bold.}\vspace{-1mm}
	\label{table1}
	\renewcommand\arraystretch{1.05}
	\begin{tabular}{c|c|c|c|c|c|c|c|c|c|c}
		\toprule
		\multirow{2}*{Methods} & \multicolumn{5}{c|}{SOTS-indoor} & \multicolumn{5}{c}{SOTS-outdoor} \\
		\cline{2-11}
		&PSNR$\uparrow$ &SSIM$\uparrow$ &NIQE$\downarrow$ &PIQE$\downarrow$ &FADE$\downarrow$ 	&PSNR$\uparrow$ &SSIM$\uparrow$ &NIQE$\downarrow$ &PIQE$\downarrow$ &FADE$\downarrow$\\
		\midrule
		AECR-Net \cite{wu2021contrastive}
		&37.17 &0.9901 &--&--&--&--&--&--&--&--\\
		PSD \cite{chen2021psd}
		&12.50 &0.7153 &4.6350&35.1740&0.6023&15.51&0.7488&3.2619&11.7166&0.7235\\
		MAXIM-2S \cite{tu2022maxim}
		&38.11&0.9908&4.2962&31.8841&0.5209&34.19&0.9846&3.0964&7.7737&0.7697\\
		Dehamer \cite{guo2022image}
		&36.63&0.9881&4.6482&31.4427&0.5020&35.18&0.9860&3.0088&9.7072&0.8318\\
		D4 \cite{yang2022_D4}
		&25.24&0.9320&4.3261&31.5310&0.7084&25.83&0.9560&2.9474&7.6678&0.7340\\
		Dehazeformer \cite{song2023vision}
		&40.05&0.9958&4.3276&32.2755&0.4987&34.95&0.9840&2.9507&7.8495&0.8093\\
		C2PNet \cite{zheng2023C2PNet}
		&42.56&0.9954&4.3978&31.7937&0.4983&36.68&0.9900&3.1437&11.7275&1.0630\\
		MB-TaylorFormer \cite{Qiu2023ICCV}
		&\bf42.63&0.9942&4.3047&31.9366&0.4959&\bf38.09&0.9910&2.9952&7.9968&0.8378\\
		MITNet \cite{shen2023MITNet}
		&40.23&0.9920&4.2529&31.3239&0.5019&35.18&0.9920&2.9454&7.8575&0.7563\\
		\textbf{Proposed}
		&42.18&\bf0.9967&\bf4.2403&\bf31.3046&\bf0.4907&36.56&\bf0.9931&\bf2.9377&\bf7.6498&\bf0.7112\\
		\bottomrule 
	\end{tabular}
\end{table*}
\begin{figure*}[htbp]
	\centering
	\includegraphics[width=0.975\linewidth]{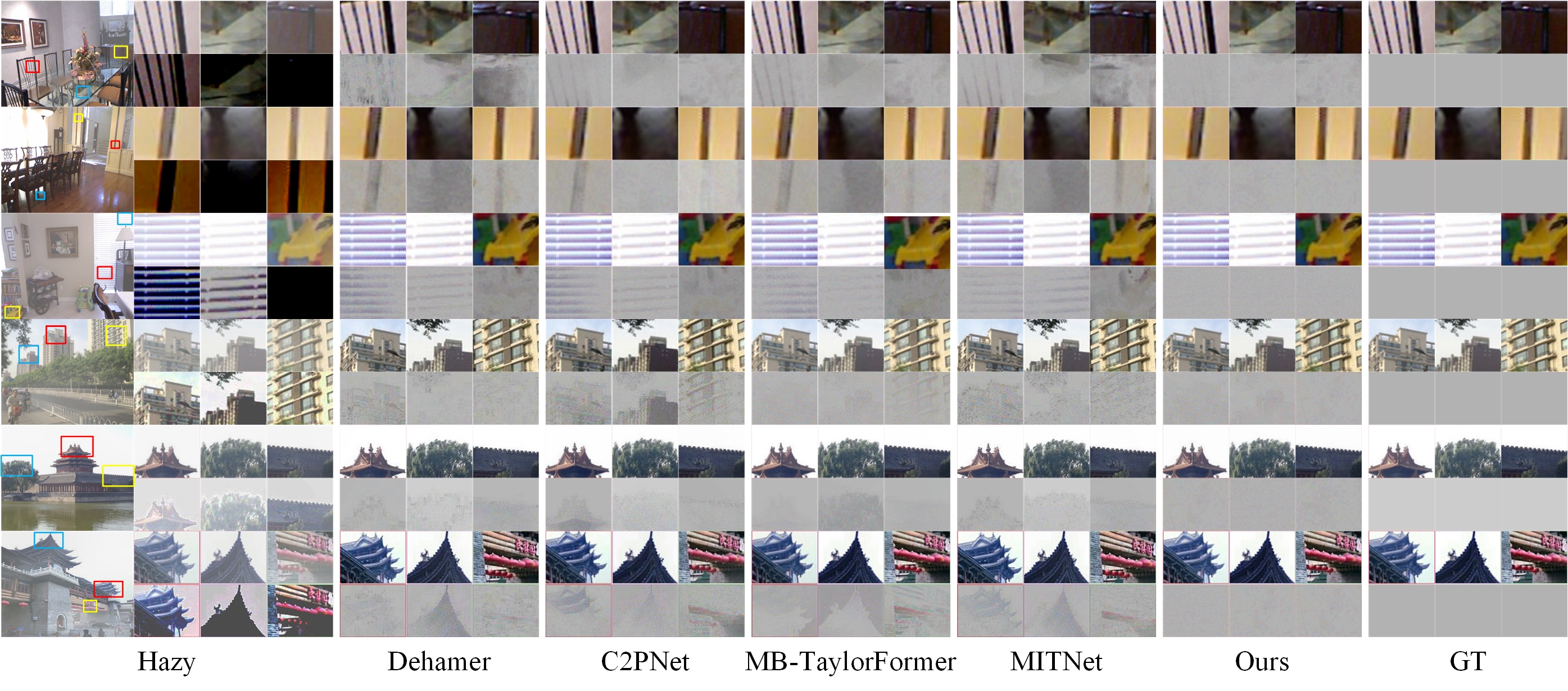}
	\caption{Visual comparisons on SOTS-indoor and SOTS-outdoor. Due to space limitations, we only show the visual effects of the images obtained by the methods with excellent performance of each year in Table~\ref{table1}. To facilitate the visual comparison, we display the visual effects of enclosed areas from the dehazed results and the differences between areas enclosed and their GTs. Less residual information in the difference map indicates better dehazing effect.} \vspace{-3mm}
	\label{Fig5}
\end{figure*}

\paragraph{Evaluation Metrics.} The proposed method is compared with the state-of-the-art deep learning-based dehazing methods. Their performance in terms of fidelity and perceptual quality is evaluated by five metrics: PSNR, SSIM, NIQE \cite{mittal2012making}, PIQE \cite{venkatanath2015blind} and FADE \cite{choi2015referenceless}.
 
\subsection{Comparison with State-of-the-arts}
\begin{figure*}[htbp]
	\centering
	\includegraphics[width=0.97\linewidth]{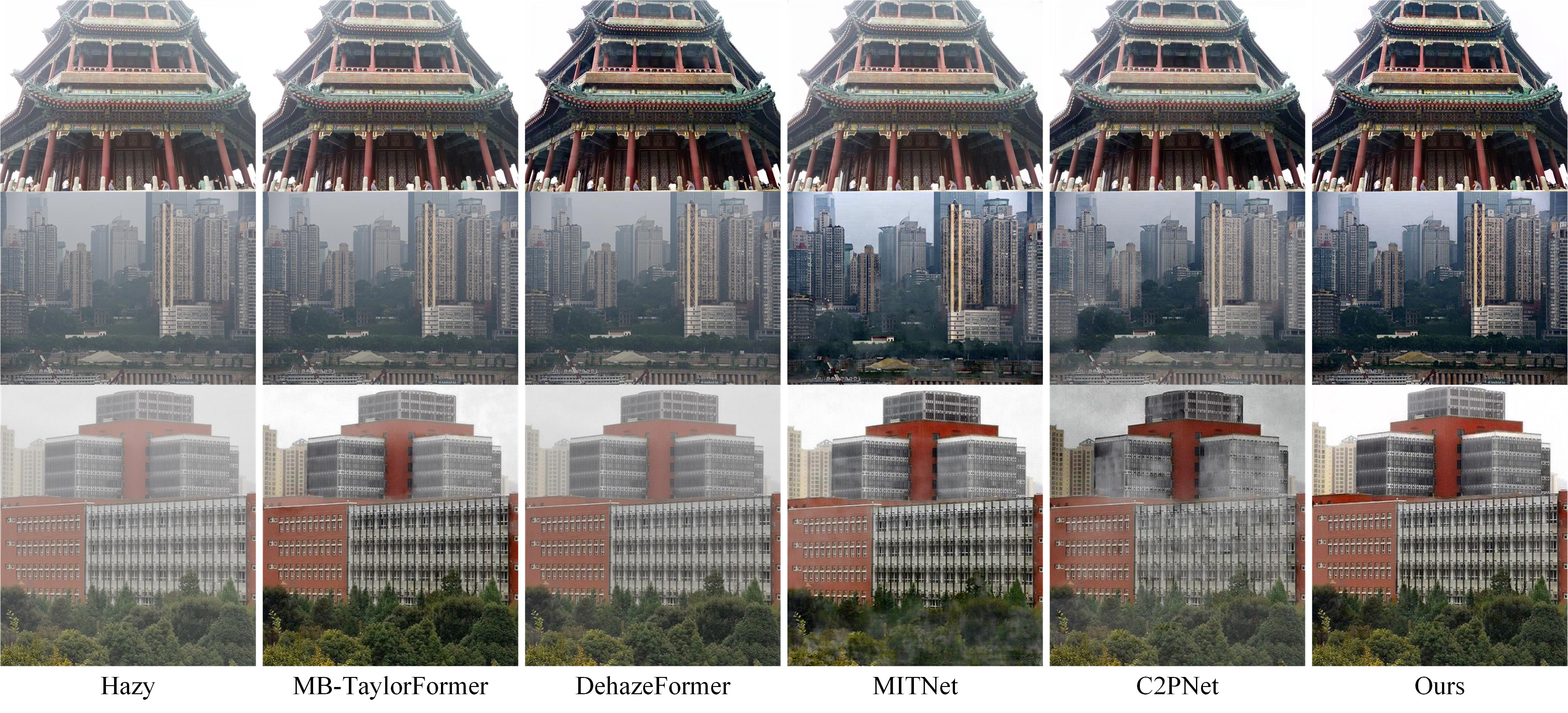}
	\caption{Visual comparisons on real-world images collected by ourselves. These images were taken casually with a mobile phone under foggy weather conditions.}
	\label{Fig6} \vspace{-3mm}
\end{figure*}

\hspace*{\parindent}Table~\ref{table1} shows the quantitative results of different dehazing methods on the SOTS indoor and outdoor datasets, respectively. The evaluation results indicate that the proposed method has achieved the best values on SSIM, NIQE, PIQE, and FADE, while also obtaining comparable scores on PSNR. Figure~\ref{Fig5} depicts the dehazing results obtained on the SOTS indoor and outdoor datasets by different methods. It can be seen from the local residual maps that the proposed method has better dehazing performance than other methods on synthetic data. Moreover, to assess the generalization of the proposed method, experiments are conducted on real-world hazy images. The experimental results are presented in Figure~\ref{Fig6}. It shows that the proposed method achieves better dehazing effect than other methods compared on real-world data.




\subsection{Ablation Study}
\hspace*{\parindent}The proposed method mainly consists of the LEGM, MFM, MSAAM, DE and DP. To verify their contribution to the full model, we conduct ablation experiments on SOTS-Indoor dataset. We exclude the MSAAM, DE, and DP from the model depicted in Figure~\ref{Fig3}, and replace the LEGM and MFM with a summation operation, then construct the Baseline. It is trained with constrains of loss in Eq.(11).
The quantitative evaluation results of the ablation experiment are listed in Table~\ref{table2}. \vspace{-5mm}

\begin{table}[htbp]\small
	\centering 
	\caption{Ablation study of each module on SOTS-Indoor dataset.} \vspace{-1mm}
	\label{table2}
	\renewcommand\arraystretch{1.05}
	\begin{tabular}{cccccc}
			\toprule
			Methods  &PSNR$\uparrow$ &SSIM$\uparrow$ &NIQE$\downarrow$ &PIQE$\downarrow$ &FADE$\downarrow$\\
			\midrule
			Baseline &35.69&0.9900&4.3974&31.9224&0.5950\\
			+LEGM &40.41&0.9947&4.2701&31.4971&0.5948\\
			+MFM &40.60&0.9957&4.2601&31.3885&0.5717\\
			+MSSAM &40.75&0.9957&4.2485&31.3657&0.5423\\
			+DE &41.55&0.9958&4.2269&31.3438&0.5018\\
			+DP&42.18&0.9967&4.2400&31.3050&0.4907\\		
			\bottomrule   \vspace{-10mm}
	\end{tabular} 
\end{table}
\begin{figure}
	\centering
	\includegraphics[width=1.0\linewidth]{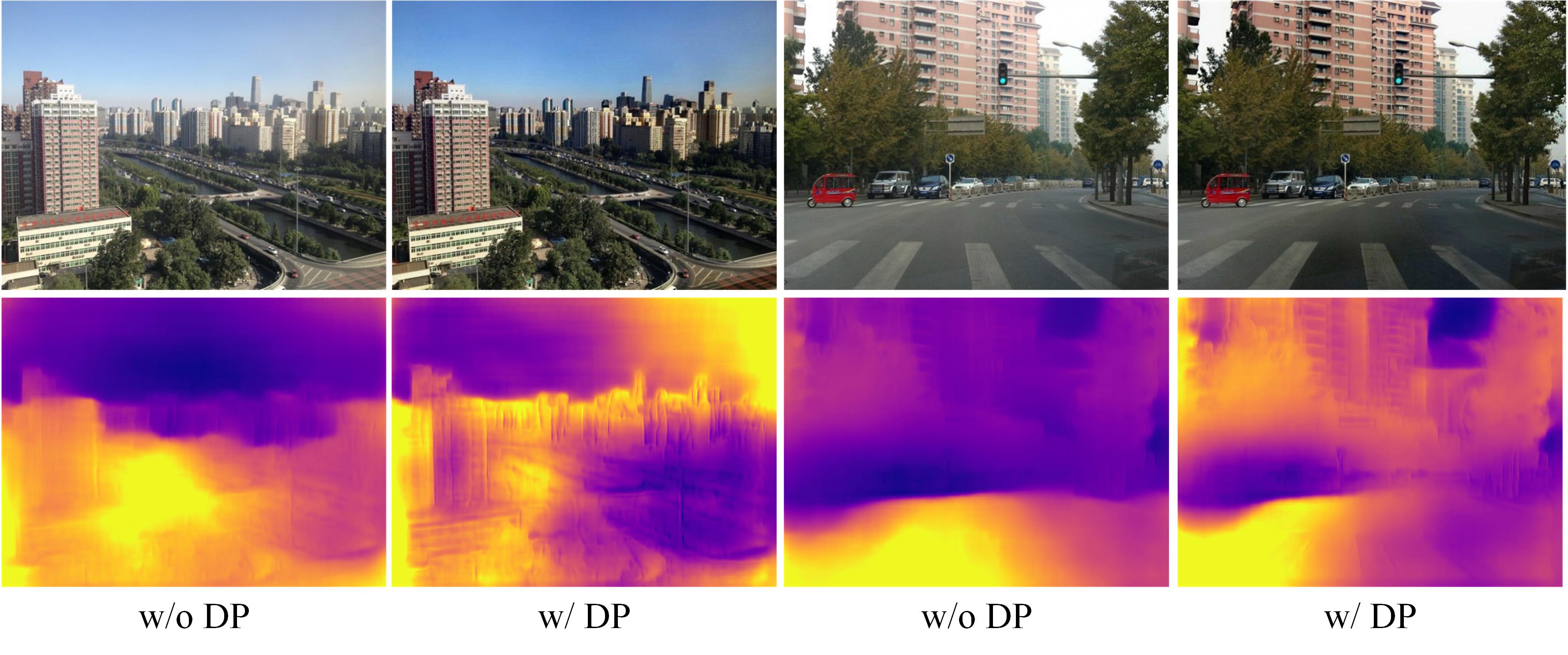}
	\caption{Depth estimation and Dehazing results of the proposed method without DP and with DP.} \vspace{-3mm}
	\label{Fig7}
\end{figure}

\paragraph{Effectiveness of LEGM.} As shown in Table~\ref{table2}, compared to the Baseline, the PSNR of the Baseline+LEGM increases by $4.72dB$. Moreover, in terms of other metrics, notable enhancements can be also observed. Those improvements can be attributed to the inherent capability of the LEGM, which integrates both global and local features. \vspace{-4mm}

\paragraph{Effectiveness of MFM.} To enhance the feature representation ability of the network, MFM facilitates the feature interaction across diverse channels by dynamically adjusting fusion weights. As listed in Table~\ref{table2}, the addition of MFM to the Baseline+LEGM improves the model performance. \vspace{-4mm}

\paragraph{Effectiveness of MSAAM.} As shown in Table~\ref{table2}, the performance of the model is further improved after MSAAM is added into Baseline+LEGM+MFM. The features of different scales aggregated through MSAAM are fused with the decoding layer features, which could alleviate the feature dilution problem in the encoding and decoding process. \vspace{-4mm}

\paragraph{Effectiveness of DE.} DE is incorporated into the Baseline+LEGM+MFM+MSAAM to verify its effectiveness. As listed in Table~\ref{table2}, the DE effectively enhances the performance of the model. This can be attributed to the fact that depth information offers crucial insights, including object depth and spatial coherence. It contributes to the preservation of image structure stability, which also proves the effectiveness of dual task mutual promotion. \vspace{-4mm}

\paragraph{Effectiveness of DP.} 
DP is added into Baseline+LEGM+MFM+MSAAM+DE to verify its effectiveness. As compared in Table~\ref{table2}, the DP facilitates the enhancement of model performance. It is discernible from Figure~\ref{Fig7} that the network with DP module acquires a noteworthy improvement in both depth estimation and dehazing results. \vspace{-1mm}

\section{Conclusion}
\hspace*{\parindent}We propose a novel approach to treat depth estimation and image dehazing as two independent tasks, and integrate them into a joint learning framework through a dual task interaction mechanism for joint optimization. The proposed difference perception in the depth maps between the dehazing result and the ideal image allows the dehazing network to focus on the sub-optimal dehazing areas to improve the network performance. Concurrently, the difference between the dehazed image and its label guides the depth estimation network to pay attention to the sub-optimal dehazing areas, thereby improving the depth prediction accuracy. In this manner, the image dehazing and depth estimation network are mutually reinforced. Experimental results show that the proposed method exhibits superior performance on both synthetic and real-world hazy images. \vspace{-2mm}

\paragraph{Acknowledgments.} This work was supported in part by the National Natural Science Foundation of China (62161015, 62276120), and the Yunnan Fundamental Research Projects (202301AV070004, 202401AS070106). 